\documentclass{article}

    \PassOptionsToPackage{numbers, compress}{natbib}

\usepackage[preprint]{neurips_2024}

\makeatletter
\renewcommand{\@noticestring}{}
\makeatother

\usepackage[hidelinks]{hyperref}

\usepackage[utf8]{inputenc} 
\usepackage[T1]{fontenc}    
\usepackage{hyperref}       
\usepackage{url}            
\usepackage{booktabs}       
\usepackage{amsfonts}       
\usepackage{nicefrac}       
\usepackage{microtype}      
\usepackage{xcolor}         

\usepackage{capt-of} 
\usepackage{amsmath}
\usepackage{graphicx}
\usepackage{subcaption}
\usepackage{colortbl}

\usepackage{algorithm}
\usepackage{algpseudocode}

\title{CoordFlow: Coordinate Flow for Pixel-wise Neural Video Representation}

\author{%
  Daniel Silver \\
  Technion, Israel Institute of Technology\\
  \texttt{silver@campus.technion.ac.il} \\
  \And
  Ron Kimmel \\
  Technion, Israel Institute of Technology\\
  \texttt{ron@cs.technion.ac.il} \\
}

\begin{document}

\maketitle

\begin{abstract}
In the field of video compression, the pursuit for better quality at lower bit rates remains a long-lasting goal. Recent developments have demonstrated the potential of Implicit Neural Representation (INR) as a promising alternative to traditional transform-based methodologies. Video INRs can be roughly divided into frame-wise and pixel-wise methods according to the structure the network outputs. While the pixel-based methods are better for upsampling and parallelization, frame-wise methods demonstrated better performance. We introduce CoordFlow, a novel pixel-wise INR for video compression. It yields state-of-the-art results compared to other pixel-wise INRs and on-par performance compared to leading frame-wise techniques. The method is based on the separation of the visual information into visually consistent layers, each represented by a dedicated network that compensates for the layer's motion. When integrated, a byproduct is an unsupervised segmentation of video sequence. Objects motion trajectories are implicitly utilized to compensate for visual-temporal redundancies. Additionally, the proposed method provides inherent video upsampling, stabilization, inpainting, and denoising capabilities.

\end{abstract}

\section{Introduction \& related works}
\label{sec:intro}
The exponential growth in digital video content, particularly in high-definition formats, has underscored the need for efficient video compression techniques. 
For instance, a 5-second clip from a 4K resolution, 120 fps video dataset, such as the UVG dataset\cite{mercat2020uvg}, requires approximately 3 GiB of uncompressed storage. 

\textbf{Classic video and image compression} standards, such as JPEGs\cite{wallace1991jpeg, skodras2001jpeg}, MPEG\cite{le1991mpeg}, and H.264\cite{wiegand2003overview}, have been used for video encoding/decoding for decades.
These methods employ transform-based approaches, utilizing the wavelet and the discrete cosine transform \cite{shapiro1993embedded, watson1994image} combined with motion compensation \cite{neff1997very}.
While these techniques have proven effective, they have reached their limitations in adaptability and performance, prompting the exploration of machine learning-based approaches. The advent of Implicit Neural Representations (INRs) now offers a promising avenue for surpassing the limitations of conventional approaches.

\textbf{Implicit Neural Representations} represents a paradigm shift in computer graphics and vision, where signals are approximated by neural networks. INRs parameterize signals as the weights of a neural network by learning a mapping from spatial coordinates to their corresponding signal values, capitalizing on the ability of neural networks to be over-fitted to their training data. The groundbreaking work of DeepSDF\cite{park2019deepsdf} and NeRF\cite{mildenhall2021nerf} demonstrated INRs' potential in 3D scenes, catalyzing a wave of research into INRs application across various domains.
Today, INRs find utility in a broad spectrum of data types susceptible to lossy compression, including images\cite{sitzmann2020implicit,chen2021learning}, videos\cite{sitzmann2020implicit,chen2021nerv,chen2022cnerv}, audio\cite{sitzmann2020implicit}, and even in neural representations of neural networks\cite{ashkenazi2022nern}. 
This wide-ranging applicability underscores the versatility and effectiveness of INRs in achieving high-fidelity signal representation.

Video INRs can be broadly classified into two main categories:

\textbf{Frame-wise INRs} methods, as introduced by NeRV\cite{chen2021nerv}, reconstruct entire frames based solely on their time index, and are more popular due to superior rate distortion (PSNR/BPP) performance. 
These methods utilize techniques such as sinusoidal positional encoding\cite{vaswani2017attention,tancik2020fourier} and transpose convolution layers in order to exploit inherent properties of natural videos. 
However, NeRV\cite{chen2021nerv} often face challenges in capturing fine-grained temporal dynamics, highlighting the need for more advanced approaches.
Notable ones are FFNeRV\cite{lee2023ffnerv}, which incorporates optical flow into the architecture in order to capitalize on the temporal redundancy, and HiNeRV\cite{kwan2024hinerv}, which uses attention, and currently set the benchmark for frame-wise INRs methods, surpassing even traditional decoders like HEVC x265 \cite{wiegand2003overview}.

\textbf{Pixel-wise INRs} operate at the pixel level, reconstructing frames pixel by pixel based on \textit{(x, y, t)} coordinates. They offer inherent upsampling capabilities, superior parallelization during inference, and are insensitive to changes in video proportions, making them more flexible and scalable in practical applications. And yet, pixel-wise networks have been somewhat neglected in the context of video compression.

A notable milestone in this domain is SIREN, as introduced by Sitzmann \textit{et al.} \cite{sitzmann2020implicit}, which utilized sinusoidal activation functions for a coordinate-based neural representation and in now widely adopted as a baseline for pixel-wise compression.
The current state-of-the-art in pixel-wise INR is NVP\cite{kim2022scalable}, which leverages learnable positional features to achieve superior performance in both training time and peak signal-to-noise ratio (PSNR). 
NVP\cite{kim2022scalable} has demonstrated its ability to outperform other methods, such as Instant-NGP\cite{caruso20233d} and FFN\cite{tancik2020fourier}, but still falls behind after NeRV which is the first and simplest Frame-wise INR .

And yet, while recent studies indicate frame-wise INRs excel in video representation, they struggle with integrating motion due to their fixed sized output and lack of sub-frame interpolation for differentiable motion modeling.

\textbf{Our Hypothesis} is that by implicitly separating video sequences into temporally coherent atlases, such as background and foreground layers, and managing each with a dedicated network, we can mitigate redundancies and achieve superior rate-distortion performance.

\textbf{Atlases} traditionally denote a compilation of images or textures into a singular, unified image, where the original signal could be reconstructed using a mapping from the atlas.
For instance, atlas images might represent the flattened and stretched exterior surface of a 3D object, for example a globe, which can then be remapped to 3D to reconstruct the object's exterior.

Neural atlases\cite{kasten2021layered}, tailored for the domain of video editing, extend this concept by representing both the images and the wrapping function through neural networks. 
An atlas-like approach to video compression, as demonstrated by Irani \textit{et al.} \cite{irani1995video}, employs homography warping to index mosaics, showcasing the potential of atlases in efficiently encoding and compressing video content.


\section{Method}

In this section, we introduce CoordFlow, a novel pixel-wise Implicit Neural Representation (INR) approach tailored for efficient video compression. CoordFlow aims to exploit the temporal redundancy inherent in natural videos by leveraging temporally coherent structures within video sequences, enabling the decomposition of these sequences into distinct layers. For each layer, a pair of dedicated neural networks is employed: one network compensates for motion within the layer by learning the spatial location of the captured structure throughout the video, while the other is an INR that encodes the layer's data.



\subsection{CoordFlow layer}

\begin{figure}[htbp]
\centering
\includegraphics[width=1\linewidth]{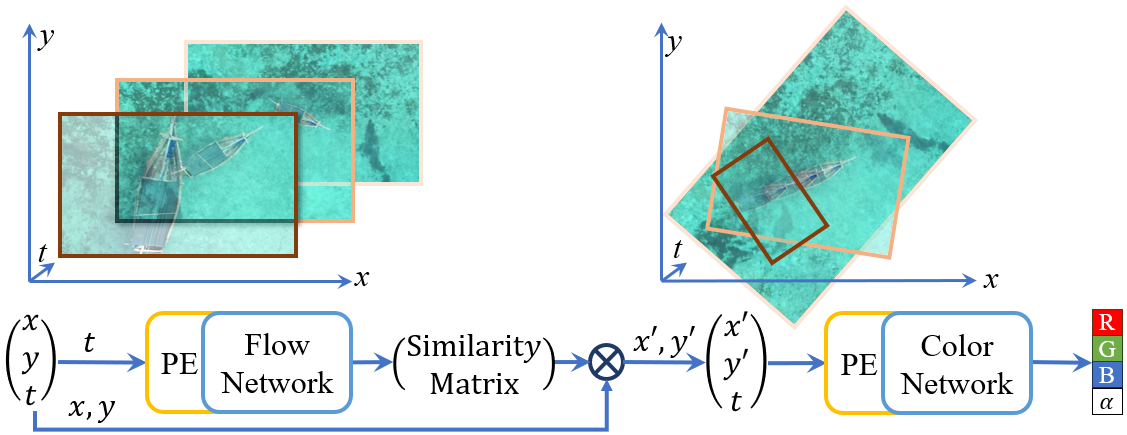}
\caption{CoordFlow layer architecture workflow. The process initiates with the input pixel coordinates \((x, y, t)\), where \(t\) undergoes positional encoding (PE) before being processed by the Flow Network. This network computes a similarity transformation to realign the spatial coordinates \((x, y)\), counteracting the motion within the video sequence, and yielding a set of transformed coordinates \((x', y', t)\). These stabilized coordinates, after positional encoding, are then inputted into the Color Network, which produces the color (RGB) and alpha (\(\alpha\)) outputs for each pixel. 
The operation of the Flow Network effectively creates a 'canonical space', in which the temporal motion is neutralized, allowing the Color Network to generate a consistent representation across time.}
\label{fig:coord_flow_single_layer}
\end{figure}
The CoordFlow layer is a specialized neural network block, whose basic architecture is tailored for video INR. 
Each CoordFlow layer can operate independently, receiving \textit{(x, y, t)} coordinates and outputting an RGB$\alpha$ value (color and opacity).
It consists of two components: a flow network and a color network. 
The flow network produces a transformation applied to the input coordinates, which are then fed into the color network to generate the layer's output.

The flow network takes only the time coordinate \textit{t} and outputs parameters for scaling \textit{s}, rotation \textit{$\theta$}, and translation \textit{$\Delta$x, $\Delta$y}. 
These parameters form a transformation matrix:

\begin{equation}
\setlength{\arraycolsep}{5pt}
F_{\text{Flow}}(t) =
\left(\begin{array}{ccc}
s \cdot \cos(\theta) & s \cdot -\sin(\theta) & \Delta x \\
s \cdot \sin(\theta) & s \cdot \cos(\theta) & \Delta y \\
\end{array}\right)
\end{equation}

This matrix is used to perform frame-wise transformations such as scaling, rotation, and translation. 
The transformed coordinates are then inputted into the color network, which outputs the value RGB$\alpha$. The inclusion of $t$ in the color network input allows it to adapt over time, compensating for the imperfections in the flow network's transformations:
\begin{eqnarray}
F_{\scriptsize\mbox{Layer}}(x,y,t) &=& F_{\scriptsize\mbox{Color}}
\begin{pmatrix}
\begin{bmatrix}
F_{\scriptsize\mbox{Flow}}(t)\times
\begin{bmatrix}
x\\
y\\ 
1\\
\end{bmatrix}\\
t
\end{bmatrix}
\end{pmatrix}
\end{eqnarray}

The CoordFlow design is motivated by the observation that natural videos commonly exhibit frame-wise movements that could be explained by similarity transformation. 
By altering the sampling area of the color network, the color network output can stay almost constant for each $(x, y)$ coordinates over different times, as can be seen in Figure \ref{fig:coord_flow_single_layer}. 
The CoordFlow approach provides a more compact representation of the video, avoiding the need to memorize pixel changes that can be explained using trivial motion.

\begin{figure}[htbp]
\centering
\includegraphics[width=1\linewidth]{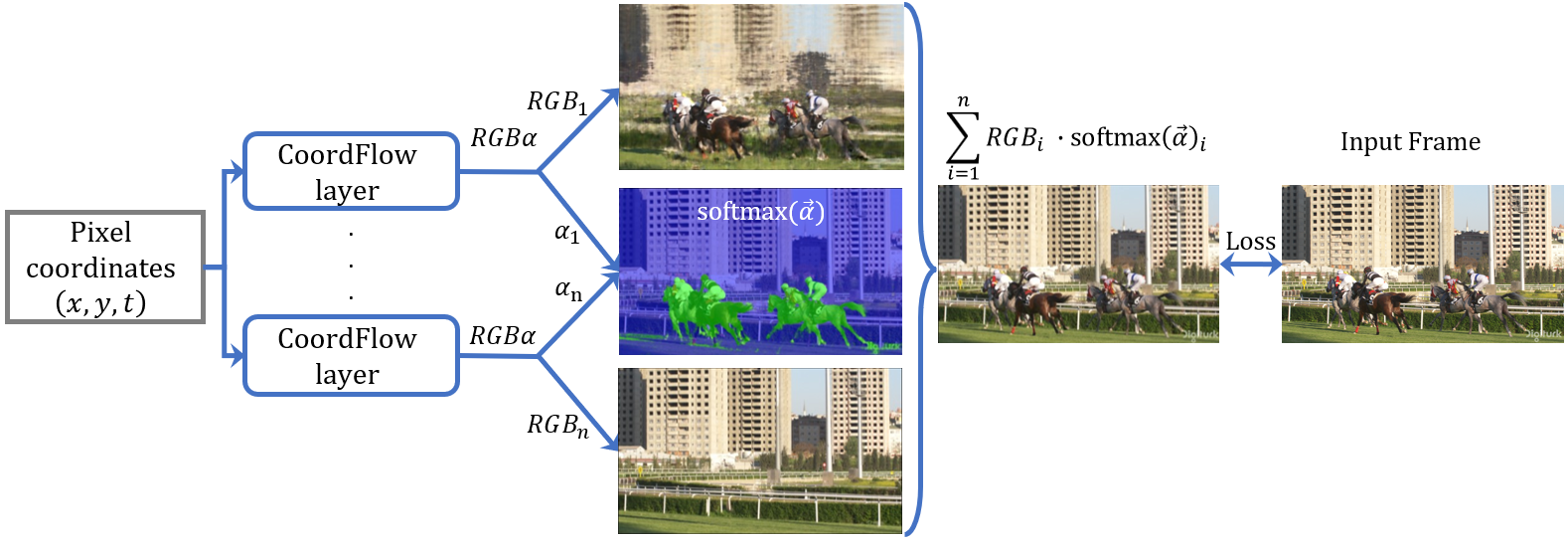}
\caption{Visualization of CoordFlow. 
The input coordinates are passed through the CoordFlow layers in parallel, outputting RGB and alpha values. 
In this example there are only two layers, and the RGB output of each layer can be seen in the middle, in addition to the softmax value of the alphas. 
The softmax map acts similarly to an attention map, and we can see the background/foreground segmentation. 
At the far right is the ground truth frame, next to the final output of the model.}
\label{fig:coord_flow_ensemble_layers}
\end{figure}

\subsection{Model architecture}

Our proposed model leverages an ensemble of CoordFlow layers. 
The quantity of these layers is a tunable hyperparameter, with potential configurations ranging from a single layer to a theoretically infinite number. 
Optimal performance is typically achieved using a two-layered structure, which aligns with the prevalent segmentation of video content into foreground and background elements.

Each CoordFlow layer is designed to process three-dimensional coordinates $(x, y, t)$, transforming them into corresponding RGB and $\alpha$ values. 
The layers operate in parallel, with each taking $(x, y, t)$ coordinates and emitting pixel-specific RGB and $\alpha$ values. 
Subsequently, a softmax function is applied to normalize the $\alpha$ values across the layers, ensuring that the layer contributions sums to one. 
The RGB values are then aggregated in a weighted sum, each value weighted by its associated softmax-normalized $\alpha$. 



To express this in a more formal manner, we denote the input coordinates $C=(x,y,t)$, and the model's output is defined by Equation~\ref{eq:model_output}, where $\overrightarrow{\alpha}$ is given by Equation~\ref{eq:alpha_vector}.

\noindent
\begin{minipage}[t]{0.6\textwidth}
\begin{equation}
F_{\text{Model}}(C) = \sum_{i=1}^{n}{ \text{softmax}(\overrightarrow{\alpha})_i \cdot F_{\text{layer[i]}}(C)[RGB]}
\label{eq:model_output}
\end{equation}
\end{minipage}%
\begin{minipage}[t]{0.4\textwidth}
\begin{equation}
\overrightarrow{\alpha} = \begin{bmatrix}
F_{\text{layer[1]}}(C)[\alpha] \\
\vdots \\
F_{\text{layer[n]}}(C)[\alpha]
\end{bmatrix}
\label{eq:alpha_vector}
\end{equation}
\end{minipage}

This architecture enables the model to autonomously segment the video into distinct regions, with each layer evolving to specialize in particular content segments. 
For instance, as depicted in Figure \ref{fig:coord_flow_ensemble_layers}, one layer may adapt to foreground elements such as moving horses, while another layer focuses on the static background. 
This unsupervised segmentation is attributable to the model's intrinsic ability of the layers to compensate for motion in a single direction, which drive each layer to focus on objects whose movement aligns with the layer's specialized direction of motion.
This method effectively reduces the memory burden on the color network by minimizing the pixel variability that the network is required to encode.



\subsection{Loss}

\textbf{Loss weighting.}
The reconstruction process is guided by a weighted loss function tailored to pixel-wise reconstruction, where traditional structural similarity indices such as SSIM\cite{wang2004image} are less applicable. 
The weighting is modulated based on the spatial-temporal frequency content of the video, aligning with the perceptual importance of different regions. 
Specifically, the weighting factor increases with the presence of high-frequency elements, as quantified by the image's Laplacian magnitude, edge prominence determined by Canny\cite{canny1986computational}\cite{kimmel2003regularized}, and temporal variance from frame to frame, with an additional constant bias to stabilize training.

\textbf{Combined L1-L2 loss.}
The composite reconstruction loss $\mathcal{L}_{\scriptsize\mbox{Combined}}$ is defined as a weighted sum of L1 (mean absolute error) and L2 (mean squared error) norms:
\begin{eqnarray}
\mathcal{L}_{\scriptsize\mbox{Combined}} = w \cdot(\|\delta\|_1 + \lambda||\delta||_2); \quad
\delta = F_{\scriptsize\mbox{Model}}(\mbox{coordinates}) - \mbox{Video(coordinates)}
\end{eqnarray}
$\delta$ is the per-pixel discrepancy between the model's output and the ground truth, $w$ is the calculated loss weighting for these coordinates, and $\lambda$ is set to $0.25$.
This combination mitigates the overemphasis on outliers by L2 norm while maintaining sensitivity to the error distribution.

\textbf{Layer-specific loss.}
To enhance the unsupervised learning of distinct video segments, we introduce a layer-specific loss $\mathcal{L}_{\scriptsize\mbox{layer}}$. 
This loss function is designed to encourage each layer within the network to focus on a subset of the video content that aligns with its learned transformation capabilities. 
The loss for each layer is computed as the product of its weighted reconstruction loss and its associated $\alpha$ values post-softmax normalization:
\begin{eqnarray}
\mathcal{L}_{\scriptsize\mbox{Layer}} &=& F_{\scriptsize\mbox{Layer}}[\alpha] \cdot \mathcal{L}_{\scriptsize\mbox{Combined}}(F_{\scriptsize\mbox{Layer}}[RGB])
\end{eqnarray}


This approach ensures that the model dynamically adjusts its internal segmentation of the video content. 
If a layer struggles to accurately represent a particular pixel, the loss function encourages it to reduce the $\alpha$ value for that pixel. 
As a result, the influence of other layers is increased, allowing the layer best suited to capture the movement direction to take the lead in its representation.

CoordFlow's learning process resembles k-means clustering convergence, where pixel associations and network transformations are iteratively refined. 
Pixels continuously shift their affiliations, while the network's directional biases adapt to align with the segment motion.

\textbf{End-to-end loss.}
This loss is a combination of both the losses mentioned: $\mathcal{L} = \mathcal{L}_{\scriptsize\mbox{Combined}} + \gamma\mathcal{L}_{\scriptsize\mbox{Layer}}$.
In the experiment section $\gamma$ was set to $0.1$.


\section{Experiments}

To evaluate the performance of our CoordFlow method, we conducted comprehensive tests for comparison with several state-of-the-art approaches, using standard benchmarks.
For pixel-wise comparisons, we included SIREN\cite{sitzmann2020implicit}, FFN\cite{tancik2020fourier}, SHACIRA\cite{girish2023shacira}, and NVP\cite{kim2022scalable}. 
In the frame-wise category, we compared against NeRV\cite{chen2021nerv}, E-NeRV\cite{chen2021nerv}, PS-NeRV\cite{bai2023ps}, FFNeRV\cite{lee2023ffnerv}, HNeRV\cite{chen2023hnerv}, and HiNeRV\cite{kwan2024hinerv}.

Our experiments were conducted on two datasets: the UVG-HD dataset\cite{mercat2020uvg}, which consists of seven 1920$\times$1080 resolution videos with 3900 frames overall, and the Boat dataset \cite{elliott2020boatvideo}, a single 1920$\times$1080 resolution video with 300 frames. 
For the UVG\cite{mercat2020uvg} dataset, we adopted the S/M/L scales used in NeRV for consistency. 
All CoordFlow models were trained for 53 epochs on a NVIDIA A40 GPU. In training learning rate was set to 0.0005 with cosine annealing scheduler and AdamW optimizer.

\begin{table}[htbp]
  \centering
  \caption{
    A comparison of CoordFlow with chosen frame-wise methods. 
    The table depicts PSNR for each method on each video from the UVG dataset, using different model sizes.
    Model results were taken from the comprehensive work of HiNeRV\cite{kwan2024hinerv} and NVP\cite{kim2022scalable}. 
    Frame wise models were tested after 300 training epochs. A more comprehensive table may be seen in the appendix.}
  {\footnotesize
  \begin{tabular}{@{}l
  @{\hspace{1em}}c
  @{\hspace{1em}}c
  @{\hspace{1em}}c
  @{\hspace{1em}}c
  @{\hspace{1em}}c
  @{\hspace{1em}}c
  @{\hspace{1em}}c
  @{\hspace{1em}}c
  @{\hspace{1em}}c
  @{\hspace{1em}}c
  @{}}
    \toprule
    Model &  Type & Size & Beauty & Bosph. & Honey. & Jockey & Ready. & Shake. & Yacht. & Avg. \\
    
    \midrule
    \multicolumn{10}{c}{Small models} \\
    \midrule
    NeRV & Frame & 3.31M & 32.83 & 32.20 & 38.15 & 30.30 & 23.62 & 33.24 & 26.43 & 30.97 \\
    E-NeRV & Frame & 3.29M & 33.13 & 33.38 & 38.87 & 30.61 & 24.53 & 34.26 & 26.87 & 31.75 \\
    HNeRV & Frame & 3.26M & 33.56 & 35.03 & 39.28 & 31.58 & 25.45 & 34.89 & 28.98 & 32.68 \\
    FFNeRV & Frame & 3.40M & 33.57 & 35.03 & 38.95 & 31.57 & 25.92 & 34.41 & 28.99 & 32.63 \\    
    \arrayrulecolor{gray!50} 
    \midrule
    \arrayrulecolor{black} 
    CoordFlow S & Pixel & \textbf{3.13M} &\textbf{33.82} & \textbf{37.76} & \textbf{39.38} & \textbf{35.09} & \textbf{29.29} & \textbf{35.42} & \textbf{30.10} & \textbf{34.40} \\ 
    
    \midrule
    \multicolumn{10}{c}{Medium models} \\
    \midrule
    
    NeRV & Frame & 6.53M & 33.67 & 34.83 & 39.00 & 33.34 & 26.03 & 34.39 & 28.23 & 32.78 \\
    E-NeRV & Frame & 6.54M & 33.97 & 35.83 & 39.75 & 33.56 & 26.94 & 35.57 & 28.79 & 33.49 \\
    HNeRV & Frame  & 6.40M & 33.99 & 36.45 & 39.56 & 33.56 & 27.38 & \textbf{35.93} & 30.48 & 33.91 \\
    FFNeRV & Frame & 6.44M & 33.98 & 36.63 & 39.58 & 33.58 & 27.39 & 35.91 & 30.51 & 33.94 \\
    
    \arrayrulecolor{gray!50} 
    \midrule
    \arrayrulecolor{black} 
    CoordFlow M & Pixel & \textbf{6.11M} & \textbf{34.05} & \textbf{38.82} & \textbf{39.59} & \textbf{36.32} & \textbf{31.12} & 35.86 & \textbf{31.58} & \textbf{35.33} \\ 

    \midrule
    \multicolumn{10}{c}{Large models} \\
    \midrule

    SIREN & Pixel & 12.60M & 27.49 & 28.31 & 31.97 & 26.58 & 19.80 & 25.23 & 23.21 & 26.09 \\
    FFN & Pixel & \textbf{10.50M} & 32.58 & 32.91 & 32.13 & 28.24 & 23.30 & 29.73 & 27.81 & 29.53 \\

    NeRV & Frame & 13.01M & 34.15 & 36.96 & 39.55 & 35.80 & 28.68 & 35.90 & 30.39 & 34.49 \\
    E-NeRV & Frame & 6.54M & 33.97 & 35.83 & 39.75 & 33.56 & 26.94 & 35.57 & 28.79 & 33.49 \\
    HNeRV & Frame & 12.87M & 34.30 & 37.96 & 39.73 & 35.47 & 29.67 & \textbf{37.16} & 32.31 & 35.23 \\
    FFNeRV & Frame & 12.66M & 33.48 & 38.48 & 39.74 & 36.72 & 30.75 & 37.08 & 32.36 & 35.63 \\

    NVP & Pixel & 136M & 34.41 & 38.40 & 37.42 & 36.97 & 32.73 & 36.86 & \textbf{33.99} & 35.83 \\
    
    \arrayrulecolor{gray!50} 
    \midrule
    \arrayrulecolor{black} 
    CoordFlow L & Pixel & 12.68M & \textbf{34.35} & \textbf{40.28} & \textbf{39.74} & \textbf{37.45} & \textbf{33.61} & 36.83 & 33.52 & \textbf{36.54} \\ 

  \bottomrule
  
  \end{tabular}
  }
  \label{tab:main}
\end{table}

\subsection{Neural representation} 
Table \ref{tab:main} presents the PSNR results for the raw (uncompressed) trained models on the UVG\cite{mercat2020uvg} dataset, and their parameters count, showcasing the performance of CoordFlow against leading methods from both the frame wise and pixel wise domains. 
As can be seen, CoordFlow secures it's competitiveness against frame wise methods and it's superiority over pixel wise ones, out preforming all methods except HiNeRV\cite{kwan2024hinerv} in terms of the average PSNR, for all size categories.
The results on the Boat video can be seen in Table \ref{tab:boat}. 
CoordFlow surpassed HiNeRV\cite{kwan2024hinerv} by more than one PSNR, while using almost half the number of parameters, thus pointing to its superiority on a certain type of natural scenery videos.
Table \ref{tab:uvg} presents the PSNR results of each raw model, no quantization, with the corresponding bit-per-pixel (BPP) value.
The results of the compared models are taken from NVP\cite{kim2022scalable} and NIRVANA\cite{maiya2023nirvana}. 
It can be seen that the small version of CoordFlow is superior to both FNN\cite{tancik2020fourier} and SIREN\cite{sitzmann2020implicit} in both BPP and PSNR, and the large version is superior to NVP\cite{kim2022scalable}. CoordFlow sets a new level for pixel-wise video representation state-of-the-art.

\begin{table}[htbp]
  \centering
  \begin{minipage}[b]{0.4\linewidth} 
    \centering
    \caption{HiNeRV and CoordFlow performance on Boat.}    
    \begin{tabular}{@{}lrr@{}}
      \toprule
      Model & Size & PSNR \\
      \midrule
      HiNeRV & 3.08M & 31.25 \\
      CoordFlow & 1.56M & 32.67 \\
      \bottomrule
    \end{tabular}
    \label{tab:boat}
  \end{minipage}
  \hfill
  \begin{minipage}[b]{0.5\linewidth} 
    \centering
    \caption{Pixel-wise methods comparison on UVG\cite{mercat2020uvg}, no quantization. Results taken from \cite{kim2022scalable,maiya2023nirvana}.}    
    \begin{tabular}{@{}lrr@{}}
      \toprule
      Model & BPP & PSNR \\
      \midrule
      CoordFlow S & 0.08 & 34.23 \\
      FNN & 0.28 & 28.18 \\
      Siren & 0.28 & 27.20 \\
      CoordFlow L & 0.33 & 36.51 \\
      NVP & 0.90 & 35.79 \\
      \bottomrule
    \end{tabular}
    \label{tab:uvg}
  \end{minipage}
\end{table}

\noindent
\begin{minipage}[h]{0.5\textwidth} 
    \subsection{Model Compression}
    The obtained neural network is compressed using weight quantization and entropy encoding. 
    The model did not go through quantization aware training and only the color network (which is approximately 98.3\% of the model) weights were quantized to 8-bit integer. 
    Quantization results can be seen in Table \ref{tab:quant}. 
    Figure \ref{fig:comprison} shows CoordFlow in comparison to other compression methods. CoordFlow is the first pixel-wise method to outperform the first frame wise INR (NeRV\cite{chen2021nerv}) and competes with standard compression methods, as H.265 \cite{sullivan2012overview}.

    
\end{minipage}
\hfill
\begin{minipage}[h]{0.45\textwidth} 
    \centering
    \includegraphics[width=0.8\linewidth]{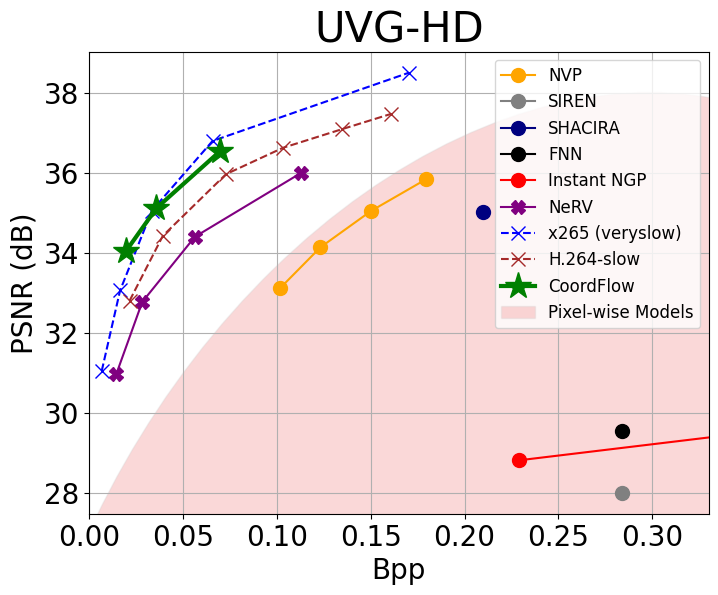}
    \captionof{figure}{Comparison to pixel-wise models, NeRV and classic methods. 
    Results taken from \cite{lee2023ffnerv, kwan2024hinerv,kim2022scalable,maiya2023nirvana}}
    \label{fig:comprison}
\end{minipage}

\begin{table}[htbp]
  \centering
  \caption{
     CoordFlow Bpp/PSNR results after quantization}
  \begin{tabular}{@{}lrrrrrrrr@{}}
    \toprule
    Model & BPP & Beauty & Bosph. & Honey. & Jockey & Ready. & Yacht. & Avg. \\    
    
    \midrule
    CoordFlow S & 0.02 & 33.75 & 37.61 & 39.24 & 34.82 & 28.86 & 30.02 & 34.05 \\ 
    CoordFlow M & 0.04 & 33.97 & 38.75 & 39.53 & 36.15 & 30.86 & 31.37 & 35.11 \\ 
    CoordFlow L & 0.07 & 34.33 & 40.05 & 39.67 & 37.13 & 33.26 & 32.70 & 36.19 \\ 

  \bottomrule
  \end{tabular}
  \label{tab:quant}
\end{table}

\subsection{Ablation study}
In this section, we examine the impact of different components within the CoordFlow architecture. Table \ref{tab:Ablation} illustrates the contribution of each module, comparing the performance of the full CoordFlow S model against its variants.


\begin{table}[htbp]
  \centering
  \caption{CoordFlow Ablation}
  \label{tab:Ablation} 
  \begin{tabular}{@{}l
  c
  @{\hspace{0.75em}}c
  @{\hspace{0.75em}}c
  @{\hspace{0.75em}}c
  @{\hspace{0.75em}}c
  @{\hspace{0.75em}}c
  @{\hspace{0.75em}}c
  @{\hspace{0.75em}}c
  @{\hspace{0.75em}}c
  @{}}
    \toprule
    Model & Size & Beauty & Bosph. & Honey. & Jockey & Ready & Shake. & Yacht. & Avg. \\
    \midrule
    CoordFlow S & 3.13M & {\textbf{33.82}} & {\textbf{37.76}} & {\textbf{39.38}} & {\textbf{35.09}} & {\textbf{29.29}} & {\textbf{35.34}} & {\textbf{30.10}} & {\textbf{34.40}} \\ 
    w/o Layers & 3.06M & 33.54 & 34.92 & 38.98 & 33.06 & 26.59 & 34.73 & 28.93 & 32.97 \\
    w/o Layers \& Flow & 3.01M & 33.52 & 34.53 & 39.16 & 32.74 & 25.82 & 34.19 & 28.80 & 32.68 \\
    \bottomrule
  \end{tabular}
\end{table}

CoordFlow S: This is our proposed architecture, consisting of two CoordFlow layers S.

w/o Layers: This variant employs a single CoordFlow layer M instead of two CoordFlow layers S.

w/o Layers \& Flow: This variant utilizes a single CoordFlow layer M using only the color network.

\section{Advanced Applications of CoordFlow}

Using CoordFlow to represent videos offers additional benefits which arise from two main characteristics of the architecture,
\begin{itemize}
\item Continuous and boundary free input space. 
\item Canonical latent space with integrated motion compensation.
\end{itemize}


\subsection{Upsampling}

CoordFlow's continuous input space allows dynamic sampling between ground truth pixel coordinates. Thus, it is possible to upsample the video in the \textit{x, y} axes for better resolution, and in the \textit{t} axis for frame interpolation to achieve higher frame rate. 

\begin{figure}[htbp]
    \centering
    \begin{minipage}[t]{0.67\textwidth}
        \centering
        \includegraphics[width=\linewidth]{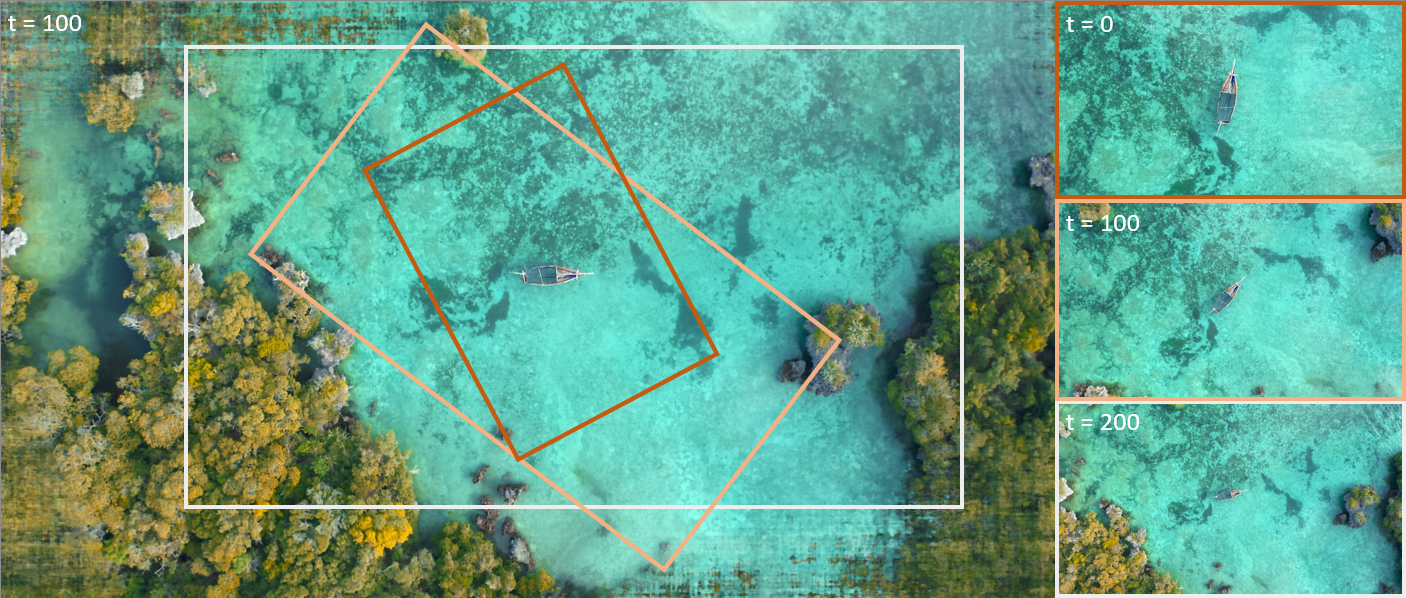}
        \caption{
        Visualization of the interaction between the flow and the color networks: The large image represents the canonical space at frame 100. This image was created by sampling the color network with \textit{x} and \textit{y} values which go out of the expected sample area of the color network for a wider view. The expected sample area for times 0, 100, and 200 are the marked boundaries. On the right are the ground truth frames, marked with the corresponding colors.
        In this example, CoordFlow refers to the video as one large image, and simply changes the sample area of the canonical space, using the flow network, in order to create the current frame.
        }
        \label{fig:Latent space}
    \end{minipage}
    \hfill
    \begin{minipage}[t]{0.29\textwidth}
        \centering
        \includegraphics[width=\linewidth]{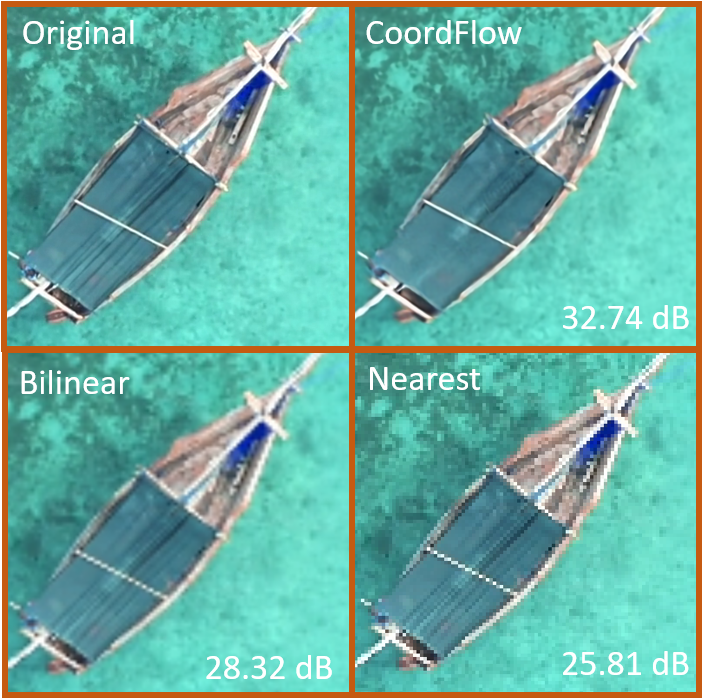}
        \caption{Upsampling of the boat video using CoordFlow (trained on every fourth pixel), bi-linear, and nearest neighbor interpolation. All methods were applied to the same down-sampled video.}
        \label{fig:Upsampling}
    \end{minipage}
\end{figure}

This inherent ability to upsample surpasses other interpolating upsampling methods like bilinear and nearest interpolation, as can be seen in Figure \ref{fig:Upsampling}. 

 CoordFlow's ability to capture motion, and by that retain a canonical space, allows for information conservation from all related video frames. As shown in Figure \ref{fig:Latent space}, CoordFlow treats videos as a single large image, retaining information across frames and incorporating multi-scale knowledge for a precise canonical representation.


\subsection{Segmentation and inpainting}
\begin{figure}[htbp]
\centering
\includegraphics[width=0.8\linewidth]{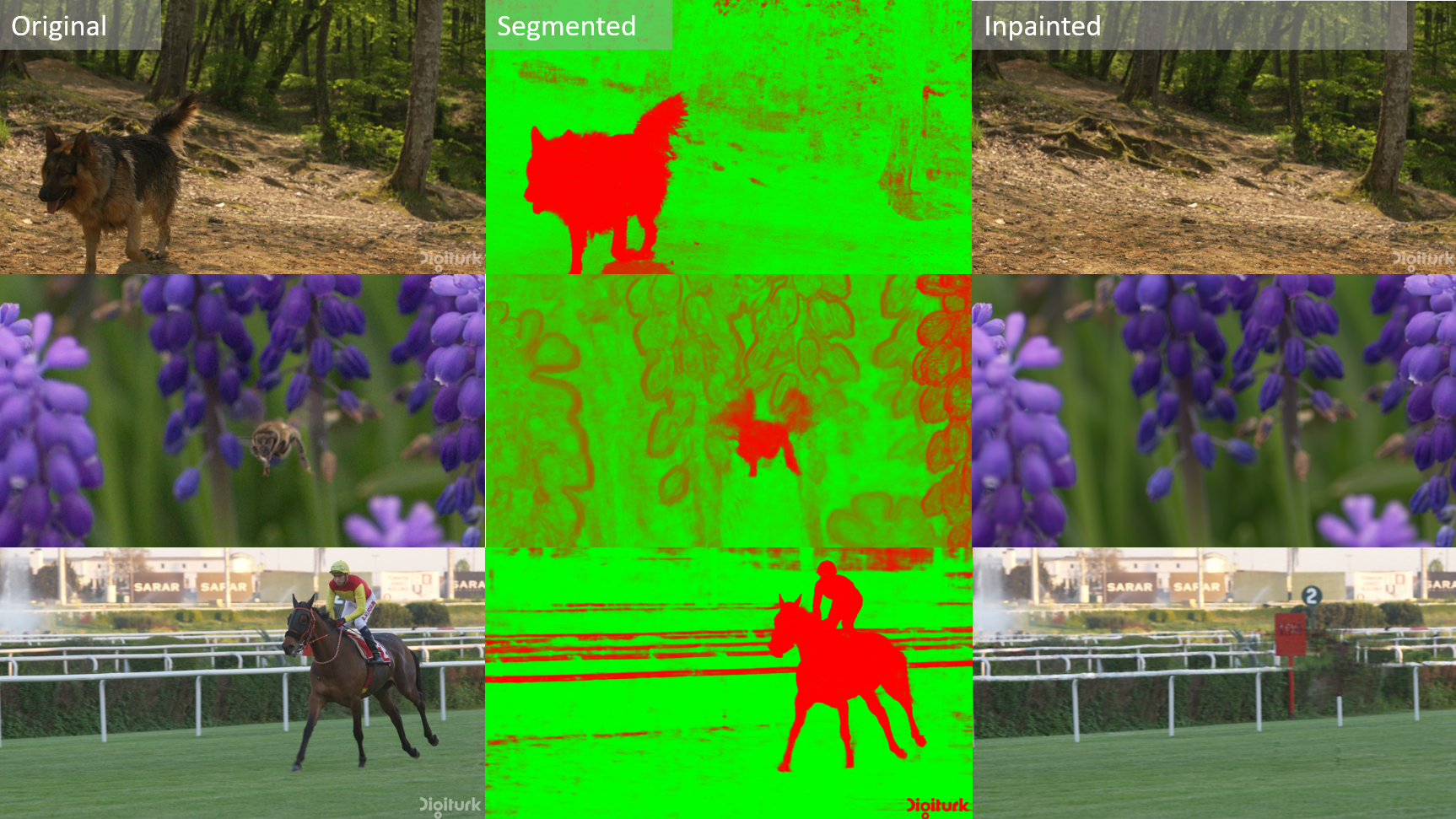}
\caption{Showcasing of the automatic inpainting capabilities: Left column - original images, middle - unsupervised segmentation, right - images without the living beings (wolf, bee, horse). 
For example, the image without the horse (bottom right) was created by the layer associated with the background (green color in the segmentation image).}
\label{fig:Inpainting}
\end{figure}

As shown in Figure \ref{fig:coord_flow_ensemble_layers}, the CoordFlow architecture is composed of a few parallel layers, each in charge of it's own part of the video.
Figure \ref{fig:Inpainting} illustrates the unsupervised segmentation created by the divergence of the layers. 
The segmentation map is the $\overrightarrow{\alpha}$ after softmax, and the inpainted image is the output of the layer associated with the background (the green color in the segmentation image). 
Other methods \cite{kim2022scalable,sitzmann2020implicit} use inpainting by masking the loss on a given segmentation, and by that use the network's interpolation ability to inpaint. 
Our method works without a need for explicit input segmentation, and due to the motion compensation can effortlessly memorize moving objects.

\subsection{Video stabilization}
Due to the motion representation explicit implementation of the network, the similarity matrix for each given time \textit{t} is extracted. 
By transforming the \textit{x, y} input by that matrix we obtain the canonical space, as can be seen in Figure \ref{fig:Latent space}. 
By smoothing the similarity matrix fluctuations over time, it is possible to achieve a more smoothly changing scene, at the cost of deviating from the sample area that the network was trained on. 
This cost is rather minimal, as CoordFlow incorporates the different frames into the canonical space. 

\subsection{Denoising}
\begin{figure}[htbp]
\centering
\includegraphics[width=1\linewidth]{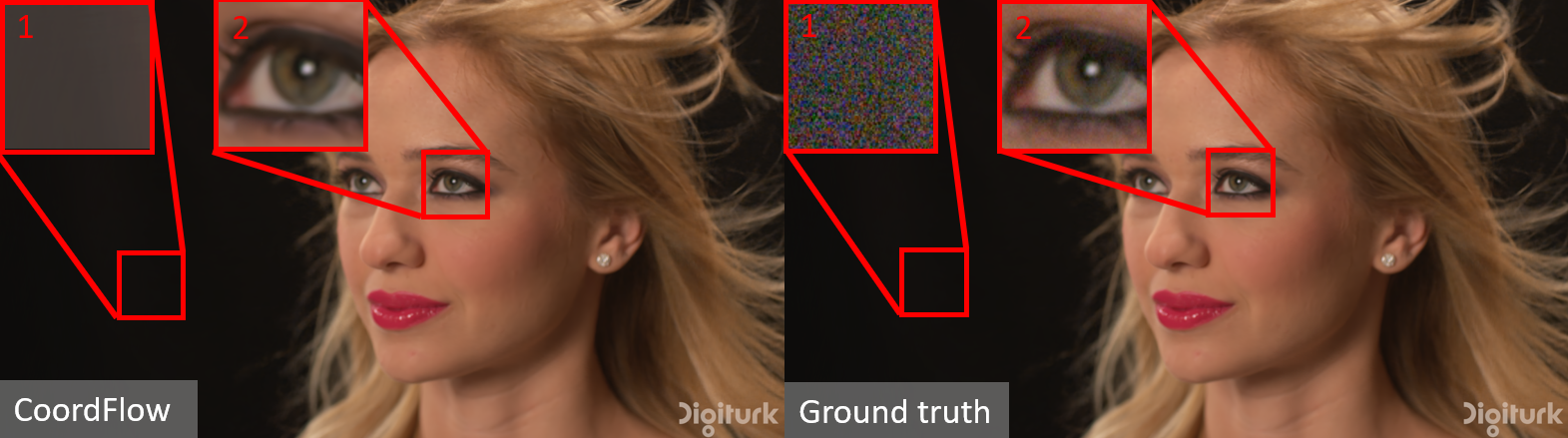}
\caption{Denoising example. Right - a frame from the Beauty video from the UVG\cite{mercat2020uvg} dataset. 
Left - the CoordFlow reconstruction (CoordFlow L from Table \ref{tab:main}). 
Square 1 and Square 2 shows that CoordFlow managed to preserve the high frequencies in the video, while successfully denoising. 
The values in square 1 were multiplied by 4 in both images for better noise visualization.}
\label{fig:denoise}
\end{figure}

Since random noise is the most difficult signal to compress, CoordFlow, by default tends to smooth out those temporal inconsistencies. 
The flow can only explain motion in the scene, and so does not contribute to the noise over-fitting. 
Thus the flow part of CoordFlow only supports non-noisy memorization, and so helps denoise the video, as can be seen in Figure \ref{fig:denoise}


\section{Conclusion}





\textbf{Summary of contributions.}
In this paper, we introduced CoordFlow, a cutting-edge pixel-wise INR framework for video compression. 
The main innovations of CoordFlow are twofold: firstly, the CoordFlow layer, a novel component adept at compensating for motion, reducing temporal redundancy; secondly, a multilayered architecture, composed of such layers, that allows for adaptive segmentation and specialized representation of different objects within a video sequence.
Our approach achieves state-of-the-art performance among pixel-wise INRs by a large margin and is the first pixel-wise method to outperform frame-wise INRs and to have comparable results with leading frame-wise and traditional techniques.
In addition, CoordFlow representation serves as a versatile tool for a suite of video processing and editing tasks \cite{rotstein2024pathways}, providing inherent capabilities for unsupervised segmentation, inpainting, upsampling, frame interpolation, stabilization, and denoising.
CoordFlow's contributions signify a leap forward in video INRs, setting a new benchmark for pixel-wise INRs and unlocking a broad spectrum of possibilities for video enhancement and manipulation.

\textbf{Limitations and challenges.}
Despite it's promising results, CoordFlow faces limitations with slow training and inference times, affecting real-time application. Although, at least the inference time can be improved through multiple GPU parallelization.

\textbf{Future Directions.}
Future research could explore CoordFlow for canonicalizing 3D objects, enhancing 3D deep learning applications. Developing more complex architectures including attention mechanisms and combination with frame-wise methods may boost performance. 
In addition improving training and inference efficiency is crucial for practical deployment.


\textbf{Closing remarks}
CoordFlow represents a significant advancement in neural video compression, offering a novel approach that leverages the temporal coherence of video data and places pixel-wise methods back in the race. 
As we continue to refine and expand its capabilities, we anticipate that CoordFlow will play a pivotal role in shaping the future of video compression and related fields.

\small
\bibliographystyle{splncs04}
\bibliography{main}

\newpage

\appendix

\begin{center}
    \LARGE \bfseries Supplemental Material
\end{center}

\section{Supplementary Videos}

\normalsize

\subsection{Canonical Space Visualization}
Link: \url{https://youtu.be/R8NU2FUctDc}

The video presents a side-by-side comparison where the ground truth video is displayed on the left. 
On the right, we visualize the latent space of the color network, with the sampling boundary delineated by red dots. This visualization was generated by sampling the color network across a constant \textit{x} and \textit{y} range, which is distinct from the sampling area dictated by the flow network (indicated by the red-dotted rectangle).

\subsection{ShakeNDry Reconstruction and Segmentation}
Link: \url{https://youtu.be/6rdDmi5DhOY}

This video showcases three elements: the original 'ShakeNDry' video on the left, the reconstruction achieved by CoordFlow in the center, and the unsupervised segmentation produced by CoordFlow on the right. 
The segmentation illustrates how CoordFlow automatically  splits the video into distinct segments, in this case background and foreground.

\subsection{Video Stabilization}
Link: \url{https://youtu.be/ITIMQCtfa54}

The video demonstrates the stabilization capabilities of CoordFlow. 
In this side-by-side comparison we present a shaky video of a scenic view on the left, while on the right, we display the stabilized video, as represented by CoordFlow, after smoothing it's scaling matrices in time.

\subsection{HoneyBee Upsampling}
Link: \url{https://youtu.be/xULGsGnRl1o}

This video showcases the upsampling capabilities of CoordFlow. In the upper right corner, you can view the original video after being downsampled, which serves as the training input for CoordFlow. The lower right section displays the reconstructed video by CoordFlow, sampled at the original video resolution. The lower left section features the results of traditional bilinear interpolation. The per-frame PSNR is indicated on each frame, allowing for a quantitative comparison of image quality across methods.

\subsection{Bee Inpainting}
Link: \url{https://youtu.be/9qNuAkryp4E}

The video demonstrates the video inpainting capabilities of CoordFlow.

\section{Investigating Pixel-wise Flow in CoordFlow Layer}
The CoordFlow layer employs a frame-wise similarity transformation to adjust input coordinates. 
When multiple layers are used, the network can capture some complex motions, depending on the number of layers. 
A single layer is limited to a single movement and is not designed for capturing intricate real-world motions but rather addresses straightforward dynamics. 
Allowing the flow network to manipulate each pixel independently, rather than limiting it to frame motion only, can enable more elaborate movement compensation. 
This raises the question of why this approach is not utilized.

\begin{figure}[htbp]
\centering
\includegraphics[width=1\linewidth]{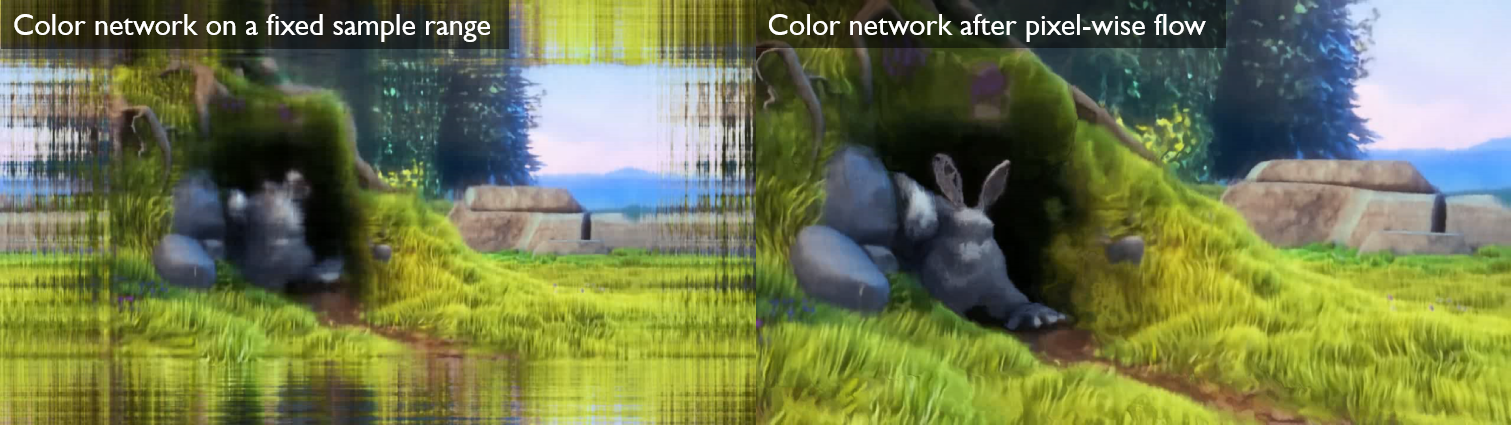}
\caption{
This figure illustrates an alternated CoordFlow layer in action, processing the ``Big Buck Bunny" sequence from scikit-video. 
The left image is generated by the color network from a fixed range of $x,y$ coordinates, revealing a distorted bunny. 
This distortion complicates the tasks of both the color network, which must memorize this warped representation, and the flow network, which has to manage the complex transformation. 
The right image shows the rectified output, exhibiting clear artifacts and high level of distortion in the bunny area.
}
\label{fig:pixel_wise_flow}
\end{figure}

Despite the ``one movement per frame" limitation, this method is preferred due to easier network training. 
Granting the flow network the freedom to manipulate each pixel independently results in sub-optimal learning outcomes, such as 'liquefied' distortions in the output image, as shown in Figure \ref{fig:pixel_wise_flow}. 
This inefficiency causes the network to expend resources correcting these distortions rather than focusing on accurate representation. 
By restricting transformations to a per-frame basis, these issues are mitigated, leading to a more stable and efficient learning process.

\section{Layer Architecture}
\begin{figure}[htbp]
    \centering
    \begin{subfigure}[b]{0.3\textwidth}
        \centering
        \includegraphics[width=\textwidth]{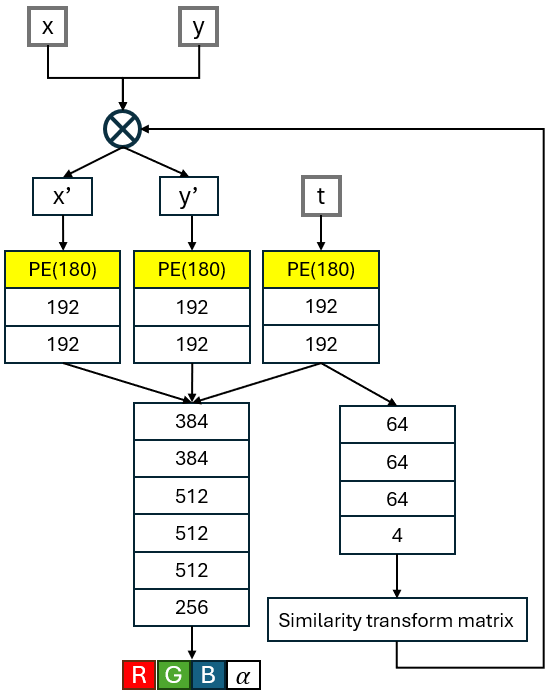}
        \caption{Small Architecture}
        \label{fig:first}
    \end{subfigure}
    \hfill
    \begin{subfigure}[b]{0.3\textwidth}
        \centering
        \includegraphics[width=\textwidth]{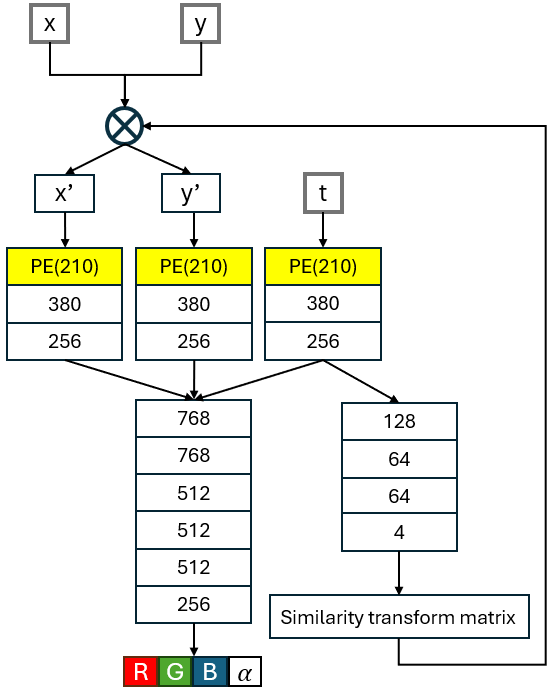}
        \caption{Medium Architecture}
        \label{fig:middle}
    \end{subfigure}
    \hfill
    \begin{subfigure}[b]{0.3\textwidth}
        \centering
        \includegraphics[width=\textwidth]{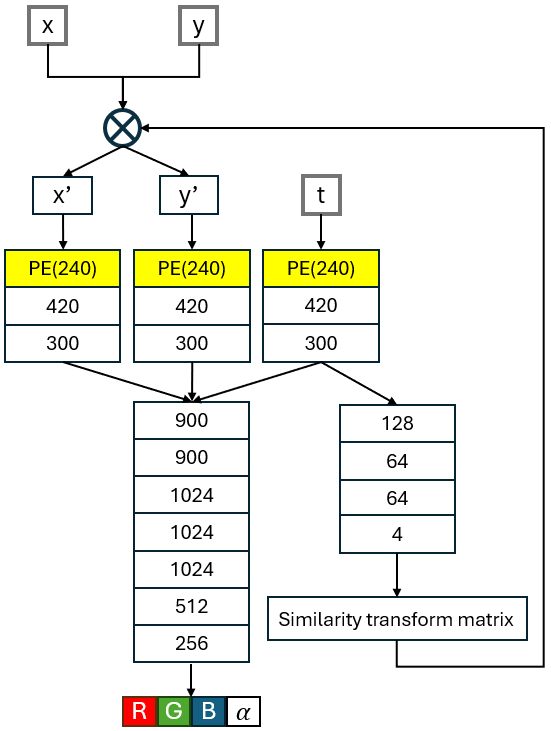}
        \caption{Large Architecture}
        \label{fig:second}
    \end{subfigure}
    \caption{
    Comparative diagrams of the different CoordFlow neural network architectures. 
    The inputs $x, y$, and $t$ processed through multiple layers, when the yellow background indicating positional encoding (PE) and the white boxes with numbers  in them represent linear layers, with the quantities specifying the number of neurons. 
    (a) The Small Architecture presents a compact design with reduced positional encoding (PE) and fewer neurons in each linear layer, tailored for lower-resolution video data. 
    (b) The Medium Architecture balances between the small and large designs, offering moderate PE sizes and neuron counts, suitable for medium-resolution video data. 
    (c) The Large Architecture includes larger PE sizes and a higher number of neurons in each layer, making it suitable for higher-resolution and more complex video data. 
    All architectures embody the method's targeted approach to capturing and compensating for motion across video frames.}
    \label{fig:three_figures}
\end{figure}

Detailed CoordFlow layer architecture for all variants can be seen in Figure\ref{fig:three_figures}.

\section{Run Times}
All run time tests were conducted on a single NVIDIA L40 GPU.

Table \ref{tab:TotalTrainingTime} describes the amount of time per training epoch as measured when training on the ShakeNDry video.

Table \ref{tab:DecodingTime} describes the amount of time for generation, as measured when training on the ShakeNDry video.

\begin{table}[htbp]
  \centering
  \begin{minipage}[t]{0.45\textwidth}
    \centering
    \caption{Training Duration per Epoch for the ShakeNDry Video}
    \begin{tabular}{@{}lrr@{}}
      \toprule
      Model Size & One Layer & Two Layers \\
      \midrule
      CoordFlow S & 6 min / epoch & 11.5 min / epoch \\
      CoordFlow M & 9 min / epoch & 22 min / epoch\\
      CoordFlow L & 16 min / epoch & 41 min / epoch\\
      \bottomrule
    \end{tabular}
    \label{tab:TotalTrainingTime}
  \end{minipage}%
  \hfill
  \begin{minipage}[t]{0.45\textwidth}
    \centering
    \caption{Decoding Time for the ShakeNDry Video}
    \begin{tabular}{@{}lrr@{}}
      \toprule
      Model Size & One Layer & Two Layers \\
      \midrule
      CoordFlow S & 02:05 (2.4 FPS) & 04:35 (1.09 FPS)\\
      CoordFlow M & 03:04 (1.6 FPS) & 06:22 (0.79 FPS)\\
      CoordFlow L & 05:52 (0.9 FPS) & 11:30 (0.43 FPS)\\
      \bottomrule
    \end{tabular}
    \label{tab:DecodingTime}
  \end{minipage}
\end{table}

\section{Additional Comparisons}

Table \ref{tab:full_table} presents a comprehensive, per video, performance comparison. The table contains table \ref{tab:main}, and also includes additional models such as PS-NeRV and HiNeRV.

\begin{table}[htbp]
  \centering
  \caption{
    A comparison of CoordFlow with SOTA frame-wise methods. 
    The table depicts PSNR for each method on each video from the UVG dataset, using tree different model sizes.
    All model results were taken from the comprehensive work of HiNeRV\cite{kwan2024hinerv}. }
  {\footnotesize
  \begin{tabular}{@{}l
  @{\hspace{1em}}c
  @{\hspace{1em}}c
  @{\hspace{1em}}c
  @{\hspace{1em}}c
  @{\hspace{1em}}c
  @{\hspace{1em}}c
  @{\hspace{1em}}c
  @{\hspace{1em}}c
  @{\hspace{1em}}c
  @{\hspace{1em}}c
  @{}}
    \toprule
    Model &  Type & Size & Beauty & Bosph. & Honey. & Jockey & Ready. & Shake. & Yacht. & Avg. \\
    
    \midrule
    
    NeRV & Frame & 3.31M & 32.83 & 32.20 & 38.15 & 30.30 & 23.62 & 33.24 & 26.43 & 30.97 \\
    E-NeRV & Frame & 3.29M & 33.13 & 33.38 & 38.87 & 30.61 & 24.53 & 34.26 & 26.87 & 31.75 \\
    PS-NeRV & Frame & 3.24M & 32.94 & 32.32 & 38.39 & 30.38 & 23.61 & 33.26 & 26.33 & 31.13 \\
    HNeRV & Frame & 3.26M & 33.56 & 35.03 & 39.28 & 31.58 & 25.45 & 34.89 & 28.98 & 32.68 \\
    FFNeRV & Frame & 3.40M & 33.57 & 35.03 & 38.95 & 31.57 & 25.92 & 34.41 & 28.99 & 32.63 \\
    HiNeRV & Frame & \textcolor{gray}{\textbf{3.19M}} & \textbf{34.08} & \textbf{38.68} & \textbf{39.71} & \textbf{36.10} & \textbf{31.53} & \textbf{35.85} & \textbf{30.95} & \textbf{35.27} \\
    
    \arrayrulecolor{gray!50} 
    \midrule
    \arrayrulecolor{black} 
    CoordFlow S & Pixel& \textbf{3.13M} & \textcolor{gray}{\textbf{33.82}} & \textcolor{gray}{\textbf{37.76}} & \textcolor{gray}{\textbf{39.38}} & \textcolor{gray}{\textbf{35.09}} & \textcolor{gray}{\textbf{29.29}} & \textcolor{gray}{\textbf{35.42}} & \textcolor{gray}{\textbf{30.10}} & \textcolor{gray}{\textbf{34.40}} \\ 
    
    \midrule
    
    NeRV & Frame & 6.53M & 33.67 & 34.83 & 39.00 & 33.34 & 26.03 & 34.39 & 28.23 & 32.78 \\
    E-NeRV & Frame & 6.54M & 33.97 & 35.83 & 39.75 & 33.56 & 26.94 & 35.57 & 28.79 & 33.49 \\
    PS-NeRV & Frame & 6.57M & 33.77 & 34.84 & 39.02 & 33.34 & 26.09 & 35.01 & 28.43 & 32.93 \\
    HNeRV & Frame  & 6.40M & 33.99 & 36.45 & 39.56 & 33.56 & 27.38 & \textcolor{gray}{\textbf{35.93}} & 30.48 & 33.91 \\
    FFNeRV & Frame & 6.44M & 33.98 & 36.63 & 39.58 & 33.58 & 27.39 & 35.91 & 30.51 & 33.94 \\
    HiNeRV & Frame & 6.49M & \textbf{34.33} & \textbf{40.37} & \textbf{39.81} & \textbf{37.93} & \textbf{34.54} & \textbf{37.04} & \textbf{32.94} & \textbf{36.71}\\

    \arrayrulecolor{gray!50} 
    \midrule
    \arrayrulecolor{black} 
    CoordFlow M & Pixel & \textbf{6.11M} & \textcolor{gray}{\textbf{34.05}} & \textcolor{gray}{\textbf{38.82}} & \textcolor{gray}{\textbf{39.59}} & \textcolor{gray}{\textbf{36.32}} & \textcolor{gray}{\textbf{31.12}} & 35.86 & \textcolor{gray}{\textbf{31.58}} & \textcolor{gray}{\textbf{35.33}} \\

    \midrule

    NeRV & Frame & 13.01M & 34.15 & 36.96 & 39.55 & 35.80 & 28.68 & 35.90 & 30.39 & 34.49 \\
    E-NeRV & Frame & 13.02M & 34.25 & 37.61 & 39.74 & 35.45 & 29.17 & 36.97 & 30.76 & 34.85 \\
    PS-NeRV & Frame & 13.07M & \textcolor{gray}{\textbf{34.50}} & 37.28 & 39.58 & 35.34 & 28.56 & 36.51 & 30.28 & 34.61 \\
    HNeRV & Frame & 12.87M & 34.30 & 37.96 & 39.73 & 35.47 & 29.67 & \textcolor{gray}{\textbf{37.16}} & 32.31 & 35.23 \\
    FFNeRV & Frame & \textbf{12.66M} & 33.48 & 38.48 & 39.74 & 36.72 & 30.75 & 37.08 & 32.36 & 35.63 \\
    HiNeRV & Frame & 12.82M & \textbf{34.66} & \textbf{41.83} & \textbf{39.95} & \textbf{39.01} & \textbf{37.32} & \textbf{38.19} & \textbf{35.20} & \textbf{38.02} \\

    FFN & Pixel & 10.50M & 32.58 & 32.91 & 32.13 & 28.24 & 23.30 & 29.73 & 27.81 & 29.53 \\
    SIREN & Pixel & 12.60M & 27.49 & 28.31 & 31.97 & 26.58 & 19.80 & 25.23 & 23.21 & 26.09 \\
    NVP & Pixel & 136M & 34.41 & 38.40 & 37.42 & 36.97 & 32.73 & 36.86 & 33.99 & 35.83 \\

    \arrayrulecolor{gray!50} 
    \midrule
    \arrayrulecolor{black} 
    CoordFlow L & Pixel & \textcolor{gray}{\textbf{12.68M}} & 34.35 & \textcolor{gray}{\textbf{40.28}} & \textcolor{gray}{\textbf{39.74}} & \textcolor{gray}{\textbf{37.45}} & \textcolor{gray}{\textbf{33.61}} & 36.83 & \textcolor{gray}{\textbf{33.52}} & \textcolor{gray}{\textbf{36.54}} \\ 

  \bottomrule
  \end{tabular}
  }
  \label{tab:full_table}
\end{table}

\section{Video Stabilization Algorithm}

\begin{algorithm}[H]
\caption{CoordFlow Video Stabilization Algorithm}
\begin{algorithmic}[1]
\Require Input video frames $V = \{f_1, f_2, \dots, f_n\}$
\Require Initialized CoordFlow model $M$
\Ensure Stabilized video frames $S = \{s_1, s_2, \dots, s_n\}$

\State Train $M$ on input video frames $V$
\State Extract frame-wise similarity transformation matrices $T = \{T_1, T_2, \dots, T_n\}$ from CoordFlow
\State Initialize smoothed transformation matrices $T' = \emptyset$

\For{each transformation matrix $T_i$ in $T$}
    \State Compute smoothed transformation matrix $T'_i$ using a temporal smoothing filter on $T_i$ and neighboring matrices
    \State Add $T'_i$ to $T'$
\EndFor

\State Initialize stabilized frames $S = \emptyset$

\For{each frame $f_i$ in $V$}
    \State Run $M$ to reconstruct the frames using the trained CoordFlow model, but instead of using $T_i$, apply the smoothed transformation matrix $T'_i$ to the input coordinates of the color network to obtain stabilized frame $s_i$
    \State Add stabilized frame $s_i$ to $S$
\EndFor \newline
\Return $S$

\end{algorithmic}
\end{algorithm}

\end{document}